\pdfoutput=1

\documentclass[11pt]{article}

\usepackage{acl2023}

\usepackage{times}
\usepackage{latexsym}

\usepackage[T1]{fontenc}

\usepackage[utf8]{inputenc}

\usepackage{microtype}

\usepackage{inconsolata}

\usepackage{graphicx} 
\usepackage{algorithm}
\usepackage{algpseudocode}
\usepackage{amsmath}
\usepackage{adjustbox}

\usepackage{enumitem}
\setlist[itemize]{}

\title{Large Language Models Meet Knowledge Graphs to\\ Answer  Factoid Questions}

\author{
\textbf{Mikhail Salnikov\textsuperscript{1,2 $^*$}}, \textbf{Hai Le}\textsuperscript{1 $^*$}, \textbf{Prateek Rajput}\textsuperscript{1}, \textbf{Irina Nikishina}\textsuperscript{5}, \\
\textbf{Pavel Braslavski}\textsuperscript{3}, \textbf{Valentin Malykh\textsuperscript{4}}, \textbf{and Alexander Panchenko\textsuperscript{1,2}} \\
\textsuperscript{1}Skolkovo Institute of Science and Technology, \textsuperscript{2}Artificial Intelligence Research Institute,\\ \textsuperscript{3}Nazarbayev University, \textsuperscript{4}ISP RAS Research Center for Trusted AI, \textsuperscript{5}Universität Hamburg
\\
\href{mailto:mikhail.salnikov@skol.tech}{\{mikhail.salnikov, hai.le, a.panchenko\}@skol.tech} 
}

\begin{document}
\maketitle
\def\thefootnote{*}\footnotetext{These authors contributed equally to this work.}\def\thefootnote{\arabic{footnote}}
\begin{abstract}
Recently, it has been shown that the incorporation of structured knowledge into Large Language Models significantly improves the results for a variety of NLP tasks. 
In this paper, we propose a method for exploring pre-trained Text-to-Text Language Models enriched with additional information from Knowledge Graphs for answering factoid questions.  More specifically, we propose an algorithm for subgraphs extraction from a Knowledge Graph based on question entities and answer candidates. Then, we procure easily interpreted information with Transformer-based models through the linearization of the extracted subgraphs. Final re-ranking of the answer candidates with the extracted information boosts \texttt{Hits@1} scores of the pre-trained text-to-text language models by $4-6\%$.
\end{abstract}

\section{Introduction}

\begin{figure}
\includegraphics[width=1.0\columnwidth]{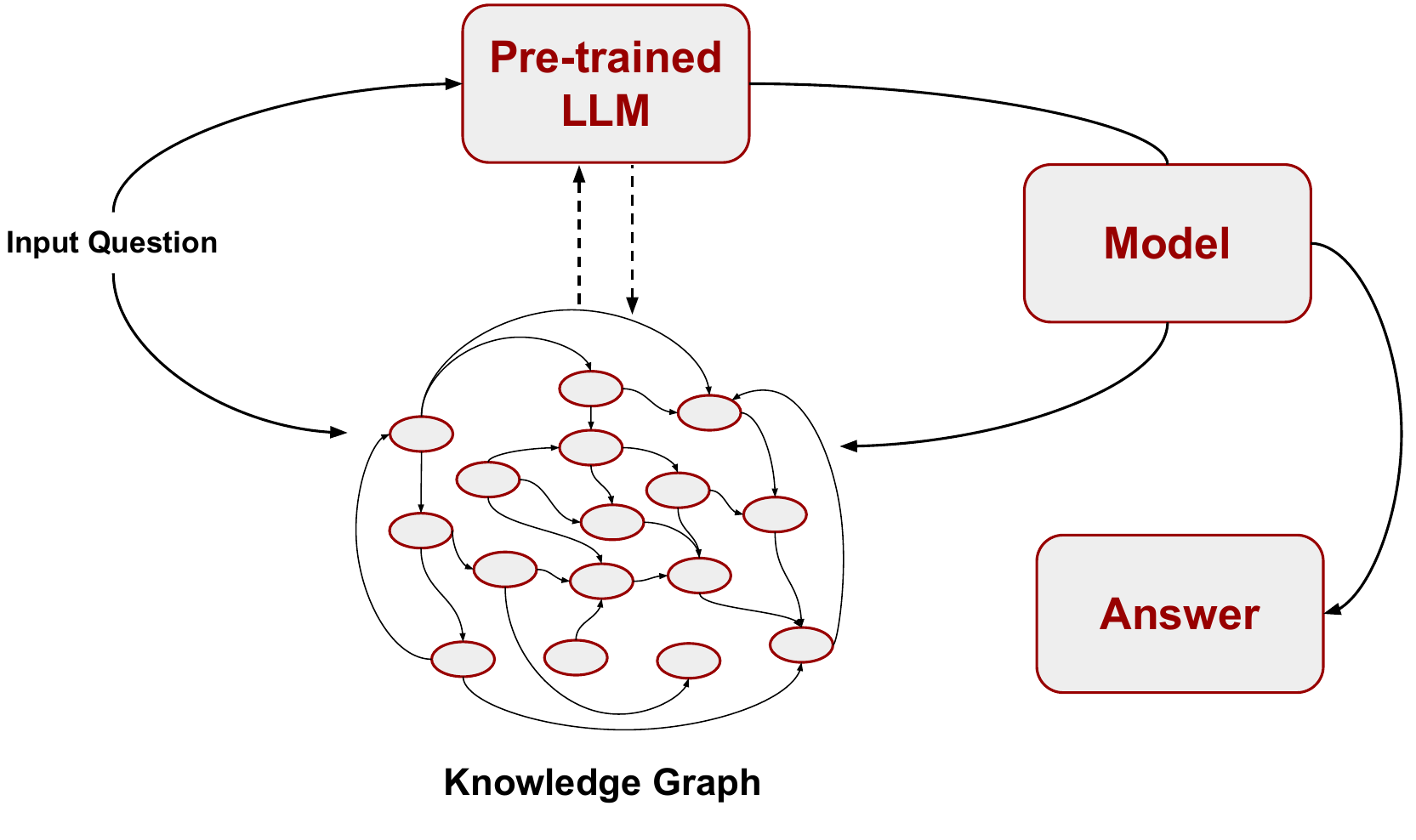}
\caption{Example of using Knowledge Graphs for answering factoid questions.}
\label{fig:sample_kg}
\end{figure}

Answering factoid questions without access to a Knowledge Graph (KG) can be challenging. The corresponding answers to these factoid questions refer to an invented or assumed statement presented as a fact, or a true but brief or trivial item of news or information. While language models can provide answers~\cite{mintaka,DBLP:conf/semweb/DubeyBA019}, the quality may not be optimal. 
Therefore, the base approach for this task relies on a structured knowledge source, such as DBPedia \cite{dbpedia}, Wikidata \cite{wikidata}, or NELL \cite{nell}. 
Given a natural language question and a corresponding KG, the goal is to predict the answer based on the analysis of the question in the context of KG. An example of how KGs could be used for answering factoid questions is presented in Figure \ref{fig:sample_kg}.

It has been shown that incorporation of the KG information into Large Language Models (LLM) significantly improves the results for various NLP tasks \cite{DBLP:conf/emnlp/ZhangLZ00H20}. At the same time, state-of-the-art Knowledge Graph Question Answering (KGQA) systems perform poorly on complex datasets \cite{mintaka}. 

\begin{figure*}[ht]
\centering{\includegraphics[width=1.0\textwidth]{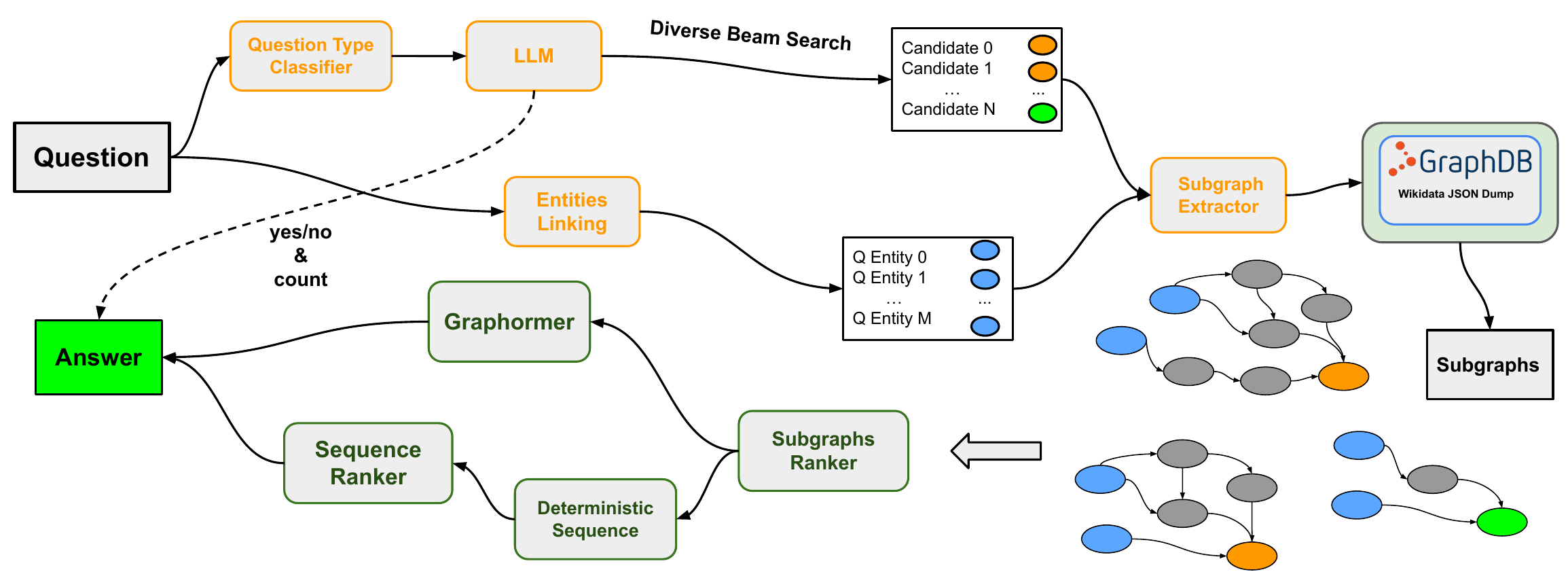}}
\caption{The proposed pipeline with subgraph extraction \& re-ranker. 
The extracted subgraphs consist of the shortest paths from the question entities to the candidates and are linearized for further ranking with Transformer Encoder.}
\label{fig:big_pipe}
\end{figure*}

In this paper, we propose a new method for KGQA based on retrieving and ranking subgraphs containing candidate answers. 
First, we generate answer candidates with an LLM and extract entities from the initial question.
Then, using a structured knowledge base, Wikidata, we construct the subgraphs 
containing those question entities and generated answer candidates. Finally, we rank these subgraphs  using the linearization of the subgraphs and raw subgraphs themselves via Transformer Encoder models. The overall pipeline can be seen in Figure \ref{fig:big_pipe}.
This approach is similar to how we browse knowledge bases such as Wikipedia or Wikidata when searching for information. Using the relevant part of the massive knowledge base graph, we ``walk" from the question entities to the potential answer and determine whether the potential answer is plausible by observing connections between them. 

Our contributions are as follows:

\begin{enumerate} 
    \item We propose a new method for KGQA by expanding and ranking answer candidates returned by a pre-trained sequence-to-sequence language model using question entities' neighbourhood.
    
    \item We present an algorithm for extraction of subgraphs corresponding to candidate answers and their subsequent ranking using Transformer Encoders. 
    

\end{enumerate}

We also publish the code, the fine-tuned models, and the subgraphs dataset on GitHub\footnote{\url{https://github.com/s-nlp/subgraph_kgqa}}.

\section{Related Work}

There exist two main approaches to KGQA: semantic parsing (translating the question to an executable logical form) and retrieval-based methods (infer answers from a Knowledge Graph). \citet{DBLP:journals/widm/ChakrabortyLMTL21,ijcai2021p0611,DBLP:journals/iet-sen/PereiraTLO22,DBLP:journals/aiopen/ZhangLFZ21} provide a detailed overview of both research directions. Therefore, we focus our attention on recent works that are 1) related to our approach or 2) trained and validated on the Mintaka dataset – our chosen complex dataset.

To start with, many retrieval KGQA methods solve the task by extracting subgraphs or neighbors based on question analysis. Then, the correct answer is chosen by searching the candidates within these subgraphs or neighbors \cite{sun-etal-2019-pullnet}. For example, \citet{WANG2023110810} present an inference chain-based model which calculates the importance of different inference chains for the question. Other models compare entity embeddings from KG with question embeddings \cite{saxena-etal-2020-improving} or entity embeddings extracted from the question \cite{razzhigaev-etal-2023-system}. Some papers present entity type prediction methods where questions are transformed into templates specifying the entity types in the input question \cite{DBLP:journals/corr/abs-1903-02419} or the answer type \cite{DBLP:conf/semweb/PerevalovB20}.

Considering the KGQA approaches to compare with, we consider the models tested on Mintaka \cite{mintaka}. The first one, \texttt{KVMemNet} \cite{miller-etal-2016-key}, operates a symbolic memory structured as key-value pairs, which gives the model a greater flexibility for encoding knowledge sources. 
\texttt{EmbedKGQA} \cite{saxena-etal-2020-improving} has three modules: Question Embedding Module, Knowledge Embedding Module, and Answer Selection Module; the latter selects the final answer based on the first two modules. Rigel~\cite{saffari-etal-2021-end} is an end-to-end QA approach which makes use of \texttt{RoBERTa} for embedding questions and performs both entity resolution and multi-hop inference. 

The approach by~\citet{10097892} also captures semantic relatedness between the questions and the paths from the knowledge base. The authors extract all paths between topic entities in similar questions and answers from the train set, and rank those paths according to their lengths. They implement an Interactive Convolutional Neural Network and score answer candidates in relation to question and paths features. Another relevant model is GreaseLM~\cite{greaselm}, where authors 
fuse encoded representations from a pre-trained Transformer Encoder and a Graph Neural Network over multiple layers of modality interaction operations. The process involves obtaining a subgraph of entities that are related to the question and then reducing it to a maximum of 200 entities. This reduction is based on a relevance score that takes into account the similarity of the embeddings. In contrast to the aforementioned approaches, our model generates candidates with LLMs and incorporates graph information into Transformers by graph linearization, which is described in \ref{sec:linearization} in more detail.



\subsection{Dataset} \label{dataset}

For our research, we focus on the Mintaka \cite{mintaka} dataset, which is a large-scale, complex and natural dataset, that can be used for end-to-end question-answering models, composed of $20,000$ question-answer pairs. This dataset is annotated with Wikidata entities and comprises 8 types of complex questions. These types include: 
\begin{itemize}
    \item \textbf{Count} (e.g., Q: How many astronauts have been elected to Congress? A: 4).
    \item \textbf{Comparative} (e.g., Q: Is Mont Blanc taller than Mount Rainier? A: Yes)
    \item \textbf{Superlative} (e.g., Q: Who was the youngest tribute in the Hunger Games? A: Rue)
    \item \textbf{Ordinal} (e.g., Q: Who was the last Ptolemaic ruler of Egypt? A: Cleopatra)
    \item \textbf{Multi-hop} (e.g., Q: Who was the quarterback of the team that won Super Bowl 50? A: Peyton Manning)
    \item \textbf{Intersection} (e.g., Q: Which movie was directed by Denis Villeneuve and stars Timothee Chalamet? A: Dune)
    \item \textbf{Difference} (e.g., Q: Which Mario Kart game did Yoshi not appear in? A: Mario Kart Live: Home Circuit)
    \item \textbf{Yes/No} (e.g., Q: Has Lady Gaga ever made a song with Ariana Grande? A: Yes)
    \item \textbf{Generic} (e.g., Q: Where was Michael Phelps born? A: Baltimore, Maryland)
\end{itemize}
Our research methodology centers around predicting entities as answers, with a particular emphasis on superlative, comparative, intersection, and multi-hop questions. However, we still compute and evaluate the results based on the complete Mintaka dataset, as our pipeline allows processing any type of questions and \texttt{yes/no} and \texttt{count} questions receive special treatment described in detail in Section \ref{sec:qtype_classifier}. 

We also compile and publish\footnote{\url{https://github.com/s-nlp/subgraph_kgqa}} the dataset of subgraphs for the whole Mintaka dataset (for train, validation, and test splits separately). Subgraphs are collected using the pipeline presented above: we generate candidate answers, we take the true answer and the entities from the question entity neighbors as candidates, and construct subgraphs with Algorithm \ref{alg:sub_extract}. As a result, we construct a ``correct'' subgraph containing the correct highlighted answer and several ``incorrect'' subgraphs from the incorrect candidate answers generated by the model. 
We present two versions of the dataset with subgraphs: with candidates generated by \texttt{T5-Large-SSM} and by \texttt{T5-XL-SSM} models.

\section{Proposed Approach}

We hypothesize that subgraphs containing paths from question entities to answer candidates provide valuable information for selecting the correct answer. Moreover, LLMs may predict an incorrect answer while still be able to generate an correct one  among top candidates. 
Thus, we generate a pool of answer candidates using a pre-trained Text-to-Text Language Model. With each answer candidate, we extract the corresponding subgraph and re-rank them based on analysis of these extracted subgraphs. Language Model generates a string as an answer, and we use Wikidata API\footnote{API URI: \url{https://www.wikidata.org/w/api.php}, method:~'wbsearchentities'} to link it to the corresponding Wikidata entity. 

Our first task (Subsections \ref{sec:candidates}-\ref{sec:subgraph}) is to construct subgraphs for each question-answer candidate pair by combining the path of the question entities to the current answer candidate. The paths from question entities to answer candidates are extracted from our chosen knowledge base, Wikidata. 
Our second objective is to rank the candidate answers using the extracted subgraphs. To do this, we compare two approaches: a Transformer-based Encoder which uses a linearized graph with a highlighted answer candidate \ref{sec:linearization}-\ref{sec:ranker} and a graph Transformer model (Graphormer) \ref{sec:graphformer}.

\begin{figure}[h]
    \centering
    \includegraphics[width=0.49\textwidth]{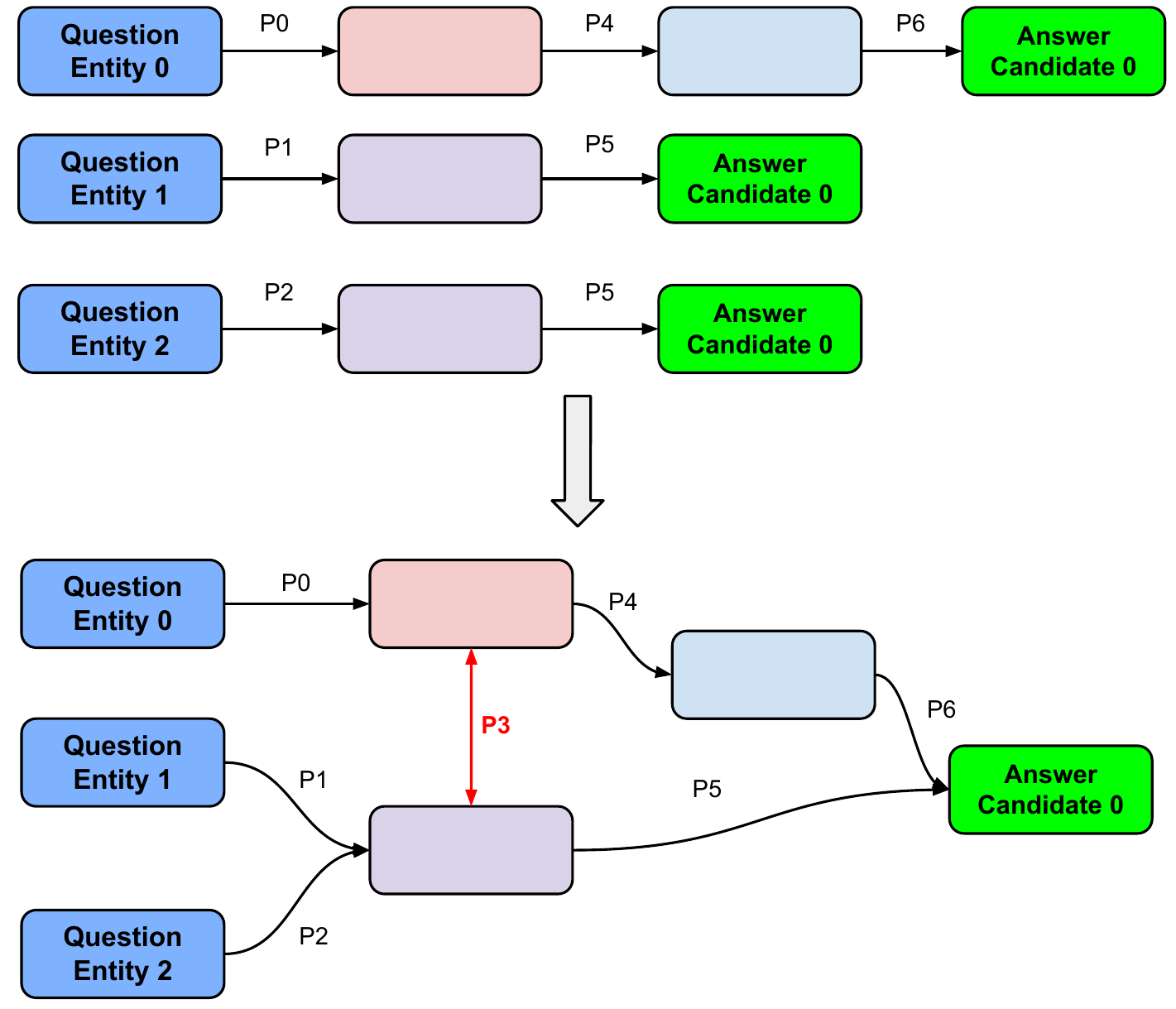}
    \caption{Subgraph Construction Algorithm – Combining extracted the shortest paths to final subgraph.}
    \label{fig:subgraph_construction_example}
\end{figure}

\subsection{Expansion of Generated Candidates} \label{sec:candidates}


Although most LLM approaches for QA, such as the one presented by \citet{mintaka}, typically use \texttt{Greed Search} and evaluate the top-1 answer, it is important to note that the correct answer may not always be the top candidate. For example, the fine-tuned T5-XL-SSM~\cite{DBLP:conf/emnlp/RobertsRS20} model achieved higher Mean Reciprocal Rank (MRR) scores for our task, indicating that re-ranking could improve the top-1 results. 
However, even when using \texttt{Classical Beam Search}, the output is often minor variations of a single sequence, which may not generate enough unique answer candidates for the Question Answering task.


To solve the problem, we apply \texttt{Diverse Beam Search} \cite{vijayakumar2018diverse}, which produces a larger number of candidates and generates them with higher variance.
\texttt{Diverse Beam Search} is formulated as follows:
\begin{equation}
    \begin{aligned}
        Y_{[t]}^g = \quad & \underset{y_1^{g}, \dots, y_{B\prime}^g \in Y_t^g} {\text{argmax}} \quad \underbrace{\sum_{b \in [B\prime]} \Theta(y_{b, [t]}^g)}_{\text{diversity penalty}} \\ 
        & + \underbrace{\sum_{h=1}^{g-1} \lambda_g \Delta(y_{b,[t]}^g, Y_{[t]}^h)}_{\text{dissimilarity term}},
    \end{aligned} 
    \label{eq::quad_func}
\end{equation}
\noindent The formula involves splitting the set of beams at time $t$ into $g$ disjointed subsets $Y_{[t]}^g$, and then selecting the candidate with the highest diversity penalty, which is calculated as the sum of a diversity penalty function $\Theta(y_{b,[t]}^g)$ over all candidates in the subset. Additionally, a dissimilarity term is included, which is calculated as the sum of a dissimilarity function $\Delta(y_{b,[t]}^g, Y_{[t]}^h)$ over all previous subsets $Y_{[t]}^h$ up to time $g-1$. The dissimilarity term is weighted by a parameter $\lambda_g$. This formula is used to optimize the selection of answer candidates in a computationally efficient manner.

\subsection{Subgraph Construction Algorithm} \label{sec:subgraph}

\begin{figure*}[h]
    \centering
    \includegraphics[width=1.0\textwidth]{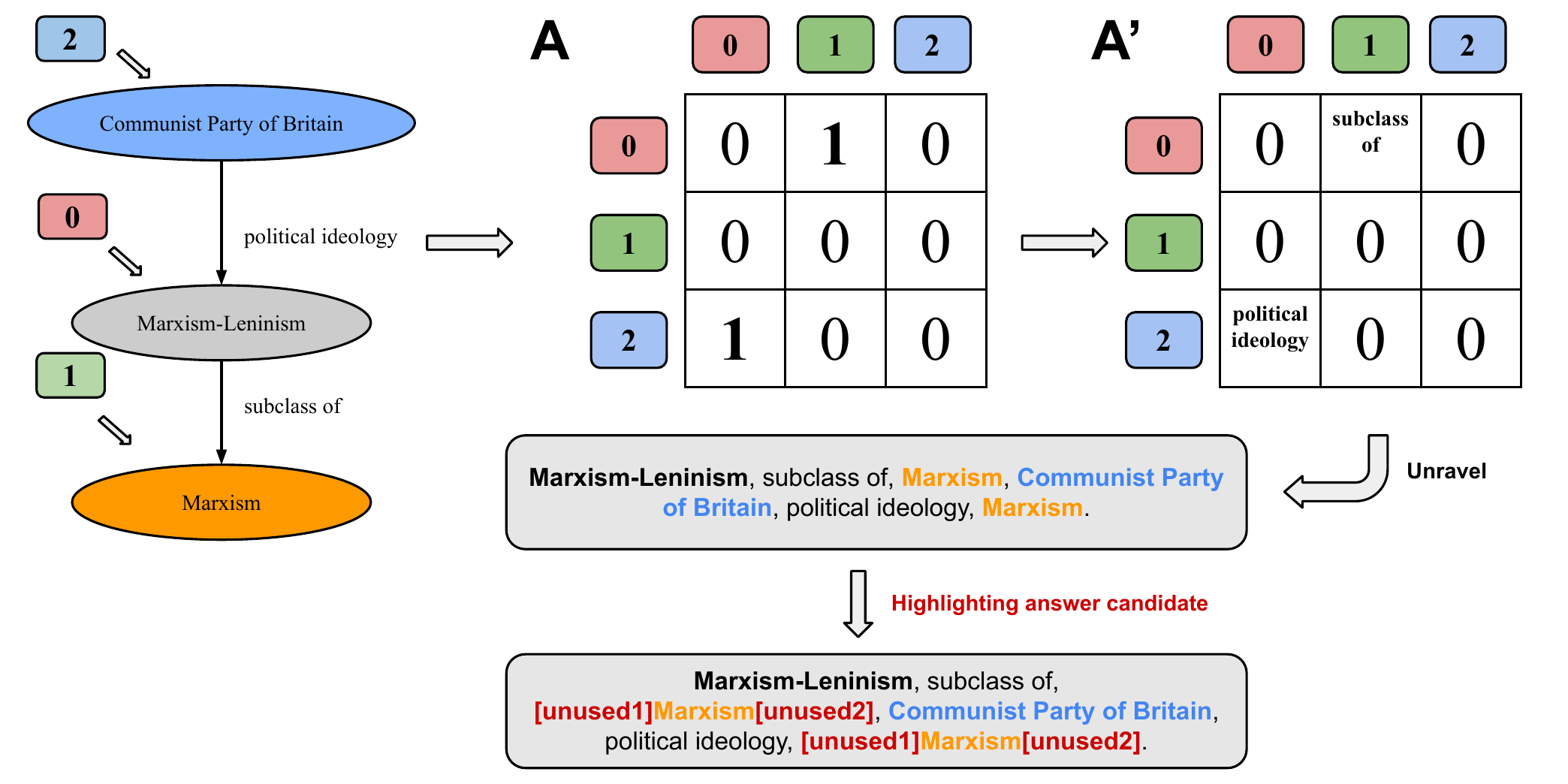}
    \caption{Subgraph Linearization. Example for question ``Which is the main ideology of the communist party of Britain?'' Entity \texttt{Q1120576} --- Communist Party of Britain taken from question.}
    \label{fig:subgraph_linearization_example}
\end{figure*}

For each question-answer candidate pair, the desired subgraph $G$ is mathematically defined as an induced subgraph of the Wikidata KG. Thus, given our shortest paths from $e_i~\rightarrow~A$, where $e_i$ - entity extracted from question and $A$ - Answer. We can use the following Algorithm~\ref{alg:sub_extract} to extract $G$. Let us define $H$ as the set of all distinct nodes within our shortest paths $P_i$. We want to preserve all edges between the nodes within $H$. For all question-answer pairs, our objective is to retain the relationship between our question entities $E$ and answer candidate entity $A_i$. The process is schematically depicted at Figure~\ref{fig:subgraph_construction_example}.

\begin{algorithm}
\caption{Subgraphs Extraction}\label{alg:sub_extract}
\begin{algorithmic}
\Require entities, candidate

\For{entity \textbf{in} entities}
\State shortest\_paths $\gets$ get shortest path from entity to candidate
\EndFor
\State H $\gets$ set of unique nodes  shortest\_paths flattened

\For{unique\_node \textbf{in} H}
\State unique\_node\_neighbor $\gets$ get neighboring nodes of unique\_node

    \For{neigh\_node \textbf{in} unique\_node\_neighbor}
        \If{neigh\_node \textbf{in} H}
            \State G $\gets$ add edge between unique\_node and neigh\_node
        \EndIf   
    \EndFor
\EndFor
\end{algorithmic}
\label{alg:sub_extract}
\end{algorithm}

\begin{algorithm}
\caption{Subgraphs to Sequence}\label{alg:sub2seq}
\begin{algorithmic}
\Require subgraph S
\State adj\_matrix $\gets$ S to adjacency matrix
\For{i \textbf{in} adj\_matrix}
    \For{j \textbf{in} i}
        \If{j \textbf{not} 0}
            \State edge\_info = edge of node $i$ and $j$
            \State final\_seq $\gets$ node $i$ label + edge\_info + node $j$ label
        \EndIf
    \EndFor
\EndFor
\end{algorithmic}
\end{algorithm}

\subsection{Graph Linearization for Sequence Ranking}
\label{sec:linearization}

To rank the extracted subgraphs, we represent the subgraphs as linearized natural language sequences, as depicted in Figure \ref{fig:subgraph_linearization_example}. Firstly, to linearize, we convert the subgraph into its binary adjacency matrix representation, $A$. 
Given $n$ nodes in the subgraph, the resulting matrix's dimension will be $n \times n$. The element $[i, j]$ of the matrix represents the existence of an edge between node with index $i$ to node with index $j$. Then, we replace the edges in the matrix with the edge label and call it $A'$.
Lastly, to produce our final sequence, we unravel $A'$ row by row and add the triple (node\_from, edge, node\_to) on to our final sequence. Algorithm \ref{alg:sub2seq} summarizes the aforementioned steps.

As a result of our analysis, we discover that 12.8\% of subgraphs in the test split of the dataset contain the correct answer but are not identified as potential candidates. To address this issue, we emphasize the answer candidate entities. To achieve this, we include special tokens before and after the answer candidate's label in the final linearized sequence, as shown in Figure~\ref{fig:subgraph_linearization_example}.

\subsection{Ranker Models}\label{sec:ranker}

With our extracted subgraphs for each respective question-answer pair, we seek to classify and re-rank to boost the original \texttt{Hits@1} scores. To achieve our objective, we test two graph-based approaches: the Transformer Encoder based approach using the linearized representation of the subgrahs and the Graphormer model using  raw subgraphs. For both methods, we train a regression model with Mean Square Error (MSE) loss. We assume that such models will be able to rank the ``correct'' subgraphs (graphs including the correct answer candidate) higher than the ``incorrect'' subgraphs (graphs including the incorrect answer candidate). 

\paragraph{Linearized Sequence Ranker}\label{sec:ranker}
For the first and main approach of our linearized sequence representation, we utilize two \texttt{BERT}-like models: \texttt{MPNet-base}\footnote{\url{https://huggingface.co/sentence-transformers/all-mpnet-base-v2}}~\cite{DBLP:conf/nips/Song0QLL20} and \texttt{DistilBERT-base}~\cite{DBLP:journals/corr/abs-1910-01108}. As input data, we provide \texttt{Question + [SEP] + Linearized subgraph} with the highlighted candidate entity using \texttt{[SEP]} tokens, as it can be seen in Figure \ref{fig:subgraph_linearization_example}. 

\paragraph{Graphormer Ranker}\label{sec:graphformer}

As an alternative to the linearized subgraphs, 
we utilize the raw subgraphs to rank our respective answer candidates. Thus, we perform graph ranking experiments using \texttt{Graphormer}~\cite{DBLP:conf/nips/YingCLZKHSL21} -- a transformer-based neural network specifically designed for graphs. We employ the \texttt{graphormer-base-pcqm4mv2}\footnote{\url{https://huggingface.co/clefourrier/graphormer-base-pcqm4mv2}} model. As input data, we encode the structural information of our subgraphs (natively in \texttt{NetworkX}\footnote{\url{https://networkx.org/}}). These structural encoding include centrality (in/out degrees), spatial (shortest path between node matrices), and edge encoding in the attention. Unlike our linearized sequences approach, the original questions are not used. 

\section{Experimental Design}
In this section, we discuss the experimental setup of our approach, which includes 1) the base Sequence-to-Sequence model used for answer candidates generation and 2) dataset used to verify the efficacy of our hypothesis and approach. 

We introduced the additional components for pieline - Question Type Classifier and Extraction of Entities from Questions. 

\subsection{Question Type Classifier}
\label{sec:qtype_classifier}

Factoid questions can sometimes have answers that cannot be linked to an entity in knowledge bases (\texttt{count} and \texttt{yes/no} questions). In order to handle this edge case, we first train a classifier to categorize questions into three types: \texttt{yes/no}, \texttt{count}, and \texttt{other}. For this classification task, we use the \texttt{MPNet} (\texttt{all-mpnet-base-v2})~\cite{DBLP:conf/nips/Song0QLL20} model with \texttt{CrossEntropy} loss. We perform $5$ epochs on the Mintaka train split, with a batch size of $32$, warm-up steps of $500$, weight decay of $0.01$, and learning rate of $0.00005$. To fit the train data to the model, we employ the HuggingFace basic Trainer and add a Weighted Random Sampler.

This pipeline demonstrates \textbf{98.29\%} balanced accuracy on Mintaka test split. With the predicted \texttt{yes/no} and \texttt{count} questions, we simply take the generated top-1 answer candidate by the pre-trained Text-to-Text Language Model. With the predicted \texttt{Other} questions, we process according to the pipeline in Figure \ref{fig:big_pipe}.

\subsection{Extraction of Entities from Questions}

With the list of answer candidates, we shift our focus to extracting the list of question entities. At this step of the pipeline, any Entity Linker like \texttt{mGENRE} \cite{decao2021multilingual} could be applied. However, in order to evaluate the approach of subgraph generation and re-ranking which are the main contribution to the task, during training and evaluation we use gold entities provided by Mintaka.

\begin{table*}
    \centering
    \begin{adjustbox}{width=0.7\textwidth,center}
    \begin{tabular}{lc}
        \hline
        \textbf{Model} & \textbf{Hits@1} \\
        \hline
        \multicolumn{2}{c}{\textbf{LANGUAGE MODELS}} \\
        \hline
        T5-Large-SSM \cite{mintaka} & 0.28\\
        T5-Large-SSM (Re-implemented) & 0.25\\
        T5-XL-SSM \cite{mintaka}& \textbf{0.38} \\
        T5-XL-SSM (Re-implemented) & \underline{0.32} \\
        ChatGPT (GPT 3.5-turbo-0301) & 0.33 \\
        \hline
        \multicolumn{2}{c}{\textbf{KGQA MODELS}} \\
        \hline
        KVMemNet \cite{miller-etal-2016-key} & 0.12 \\
        EmbedKGQA \cite{saxena-etal-2020-improving} & 0.18 \\
        Rigel \cite{saffari-etal-2021-end} & 0.20\\
        \hline
        \multicolumn{2}{c}{\textbf{OUR APPROACH WITH SUBGRAPHS AND RE-RANKING}} \\
        \hline
        T5-Large-SSM (Re-implemented) + Linearization + MPNet & 0.29\\
        T5-Large-SSM (Re-implemented) + Linearization + DistilBERT & 0.27 \\
        T5-Large-SSM (Re-implemented) + Graphormer & 0.25 \\
        T5-XL-SSM (Re-implemented) + Linearization + MPNet & \textbf{\underline{0.38}} \\
        T5-XL-SSM (Re-implemented) + Linearization + DistilBERT & 0.37 \\
        T5-XL-SSM (Re-implemented) + Graphormer & 0.32 \\
        
        \hline
    \end{tabular}
    \end{adjustbox}
    \caption{\texttt{Hits@1} for MINTAKA. A comparison between our methodology and SOTA. We show an improvement over the original language model, \texttt{Hits@1} from 0.32 to 0.38.}
    \label{tab:results}
\end{table*}

\subsection{Base Model}


As the first step of our approach, we fine-tune the \texttt{T5-SSM} models \cite{DBLP:conf/emnlp/RobertsRS20} (Large and XL) on English questions for 10,000 steps, following \citet{mintaka}. This model was reported as the state-of-the-art for the Mintaka dataset. Although we attempted to adhere to the \citet{mintaka} model's fine-tuning protocol, there were certain aspects that were overlooked, which makes it unfeasible to replicate the outcome. Despite our efforts, we were unable to replicate the results as claimed in \cite{mintaka}. Thus, in the final Table \ref{tab:results}, we provide their results and ours as \textit{re-implemented}. We used Hits@1 metric for comparing our approach with others because it is a famous metric for KGQA task and many other authors~\cite{mintaka, wikidata-benchmark} used only this one metric, making it easier to compare results. Additionally, we developed a KGQA system and were interested in getting the final answer, not a list of candidates that include the answer on some position. Therefore, the primary contribution of our work is the enhancement in quality compared to our \textit{re-implemented} model.

With the best fine-tuned \texttt{T5-SSM-XL}~\cite{DBLP:conf/emnlp/RobertsRS20} model, we generate our answer candidates pool for subgraph extraction and ranking. However, it can be replaced with any other model for candidates generation.

\section{Results and Discussion}

In the following section,  we present the results for the proposed approach in comparison with several baselines. Table \ref{tab:results} shows that the results for the re-implemented \texttt{T5}-models are significantly improved with our suggested approach: 4\% for the large model and 6\% for the XL model. We also note the increase in quality can be achieved with any Encoder, however, \texttt{MPNet} performs slightly better than \texttt{DistilBERT-base}. As our re-implemented \texttt{T5}-models results are significantly lower than presented in the paper (\texttt{Hits@1} is equal to $0.25$ and $0.32$ for large and XL models), we do not outperform the reported state-of-the-art, but perform on par with it. Application of our approach to the version trained by \citet{mintaka} should boost the scores even more.

Our second approach using \texttt{Graphormer} ranking with solely the raw subgraphs does not improve the results of an LLM. We hypothesize that re-ranking only operates on structural encodings of our subgraphs, which results in a subpar performance. Without  information from the question and a potential answer, the subgraphs cannot provide a complete and coherent representation of the question-answer pair.

Additionally, we evaluate another popular large language model --- \texttt{ChatGPT} 
with the following prompt before each question: \texttt{``Answer as briefly as possible. The answer should be 'Yes', 'No' or a number if I am asking for a quantity of something, if possible, otherwise just a few words.''} The result of \texttt{ChatGPT} can be comparable to our fine-tuned \texttt{T5-XL-SSM}, with \texttt{Hits@1} scores of $0.33$ and $0.32$ respectively. However, they significantly lag behind our proposed approach, as shown in the final Table \ref{tab:results}.

\subsection{Question Type Analysis}

In order to understand the efficacy of the proposed approach, we calculate scores for each question type in the dataset, which were described in Section \ref{dataset}. The results in Table \ref{tab:reranking_types} are based on the best-performing base model and re-ranker (\texttt{T5-XL-SMM} + \texttt{T5-Large-SSM (Re-implemented) + Linearization + \texttt{MPNet}}). We exclude questions that do not have answers with the corresponding Wikidata entity (\texttt{yes/no} and \texttt{count}), as the precision scores remain the same for these question types. For all other \texttt{ComplexityType} types of this dataset, our re-ranker successfully bolsters the \texttt{Hits@1} scores.

\begin{table}[H]
    \centering
    \begin{adjustbox}{width=0.49\textwidth,center}
    \begin{tabular}{lccc}
        \hline
        \textbf{ComplexityType} & \textbf{Original}  & \textbf{Re-ranked}  & \textbf{Hits@All} \\
        \hline
        Intersection & 0.36 & 0.53 & 0.68 
        \\
        Count & 0.25 & 0.25 & 0.94 
        \\
        Comparative & 0.50 & 0.55 & 0.96 
        \\
        Yesno & 0.62 & 0.62 & 1.00 
        \\
        Generic & 0.34 & 0.35 & 0.65 
        \\
        Ordinal & 0.21 & 0.22 & 0.59 
        \\
        Multihop & 0.14 & 0.18 & 0.45 
        \\
        Difference & 0.14 & 0.36 & 0.45 
        \\
        Superlative & 0.28 & 0.41 & 0.55 
        \\
        \hline
        All & 0.32 & 0.38 & 0.69\\
       \hline
    \end{tabular}
    \end{adjustbox}
    \caption{\texttt{Hits@1} Results based on \texttt{ComplexityType}. \texttt{Hits@All} --- upper bound.}
    \label{tab:reranking_types}
\end{table}

Upon further inspection, the largest increase in accuracy (from Hit@1 0.32 to 0.38) is for the \texttt{Intersection} type of questions (i.e. \textit{``What game was released by Impressions Games and is an expansion to Pharaoh?''}). We hypothesize that the answer of these \texttt{Intersection} question is connected to all question entities. Thus, the extracted subgraphs contain more meaningful information, representative of the relationship between our question answer pair. Additionally, \texttt{Multi-hop} and \texttt{Difference} questions also display a significant increase in scores. Despite having a minute boost in other question types, the proposed approach still aids in performance in all categories. Overall, one may argue that the re-ranker performs better with complex questions.


\subsection{Extracted Subgraphs Analysis}
As shown previously, our proposed approach generally improves \texttt{Hits@1} scores for all question types. We hypothesize that geometry of the extracted subgraphs aided in the improved \texttt{Hits@1} scores. Thus, to further understand the reasoning behind this performance boost, we examine the difference between the extracted subgraphs of the correct and incorrect question-answer pairs. We assume  that ``incorrect'' subgraphs tend to be denser (more nodes and edges); while the ``correct'' subgraphs tend to be sparser (fewer nodes and edges). We also assume that this information could be useful for our \texttt{Graphomer} ranking approach. Despite having no information regarding the question, we hypothesize that our pipeline picked up these differences between the ``correct'' vs. ``incorrect'' subgraphs via the structural encodings. To verify this hypothesis, we collect different graph metrics for further analysis. 

\begin{table}[!htb]
    \centering
    \begin{adjustbox}{width=0.49\textwidth,center}
    \begin{tabular}{lrr}
        \hline
        \textbf{Complexity Metrics} & \begin{tabular}[c]{@{}c@{}}\textbf{``Correct''}\\ \textbf{Subgraphs}\end{tabular}  & \begin{tabular}[c]{@{}c@{}}\textbf{``Incorrect''}\\ \textbf{Subgraphs}\end{tabular}  \\
        \hline
        Number of Nodes & 2.98 & 3.14 \\
        Number of Edges & 3.31 & 3.64 \\
        Density & 0.61 & 0.63 \\
        Number of Simple Cycles & 1.04 & 1.18 \\
        Number of Bridges & 2.96 & 3.10 \\
        \hline
    \end{tabular}
    \end{adjustbox}
    \caption{Average graph statistics for the generated subgraphs: for the ``correct'' and ``incorrect'' subgraphs.}
    \label{tab:subgraph_inspect}
\end{table} 

Table \ref{tab:subgraph_inspect} displays the average of the number of nodes, edges, density, simple cycles (elementary circuits), and bridges (isthmus) for the ``incorrect'' and ``correct'' subgraphs. We analyze these metrics on our subgraph dataset of 13,491 ``correct'' subgraphs and 94,615 ``incorrect'' subgraphs; which excludes \texttt{yes/no} and \texttt{count} question types. 

The above table confirms our hypothesis that ``incorrect'' subgraphs have higher density, larger number of nodes, edges, simple cycles and bridges. This might partially explain why the ranker can differentiate the ``correct'' subgraphs from the ``incorrect'' subgraphs. 
Overall, we hypothesize that the geometry of the subgraphs themselves contributes to the boost of our final \texttt{Hits@1} score. 

\subsection{Ablation study}

In this section, our objective is to determine the most crucial component of the proposed ranking model. To achieve this, we employ the best performing model and train and evaluate it, removing certain parts of the pipeline. 

First, we disable the highlighting feature and observe that the \texttt{Hits@1} score on the Mintaka test part drops from $0.38$ to $0.36$.
Second, we examine the significance of the proposed subgraph extraction approach. We train another ranking model that ranks candidates without subgraphs. As input, we provide only a question and a candidate label to the \texttt{MPNet} ranker model in the following format: $question + [SEP] + candidate$. The results show a \texttt{Hits@1} score of $0.33$, which clearly indicates that the large contribution of the subgraph information to the final score.

All in all, eliminating both parts of the input for the ranking model drastically decreases the results. Candidate highlighting as well as subgraphs are necessary for choosing the correct candidate answer.   

\subsection{Limitations}

The main limitation of the proposed system is that it is tested on the English language only. Mintaka possesses suites in other languages and our approach should definitely be evaluated on them. Moreover, while Mintaka has proved to be a new challenging dataset with various types of questions, it would be interesting to test our approach on other datasets, such as LC-QuAD~2.0~\cite{DBLP:conf/semweb/DubeyBA019} and RuBQ~2.0~\cite{DBLP:conf/esws/RybinKEB21}. Furthermore, we have not proofed the full pipeline performance as we do not embed an Entity Linker and test our model on gold question entities and have not tested other Generative Transformers.

In terms of computational efficiency, communication with a Knowledge Graph can be a bottleneck, as it might be time-consuming for generating subgraphs from a KB for all 200 answer candidates.

\section{Conclusion}

To sum up, in this paper we have proposed an approach for improving the output of LLMs for Question Answering using additional information from Knowledge Graphs. We have improved \texttt{Hits@1} by $4$ to $6$\% by extracting subgraphs relevant to the input question entities and the predicted answer candidates; and further ranking the answer candidates by the extracted subgraphs. Our result analysis shows that the suggested solution improves scores for the \texttt{Intersection} questions and has almost no effect on \texttt{Comparative} questions. The ablation study proves the efficiency of each pipeline step. As future work, we plan to extend our approach to other languages and test the full pipeline with entity linker and other generative Transformer models. 

\section*{Acknowledgements}
The work of Irina Nikishina was supported by the DFG through the project ``ACQuA: Answering Comparative Questions with Arguments'' (grants BI 1544/7- 1 and HA 5851/2- 1) as part of the priority program ``RATIO: Robust Argumentation Machines'' (SPP 1999). 
The work of Valentin Malykh was supported by a
grant for research centers in the field of artificial
intelligence, provided by the Analytical Center for
the Government of the Russian Federation in accordance with the subsidy agreement (agreement identifier 000000D730321P5Q0002) and the agreement
with the Ivannikov Institute for System Programming of the Russian Academy of Sciences dated
November 2, 2021 No. 70-2021-00142.

\bibliography{anthology-2,custom}
\bibliographystyle{acl_natbib}




\end{document}